\documentclass[10pt,twocolumn,letterpaper]{article}

\usepackage{cvpr}              

\usepackage{xcolor,colortbl}
\usepackage{url}            
\usepackage{booktabs}       
\usepackage{amsfonts}       
\usepackage{nicefrac}       
\usepackage{microtype}      
\usepackage{amsmath}
\usepackage{algorithm}
\usepackage{algorithmic}
\usepackage{float}
\usepackage{multirow, adjustbox}
\usepackage{booktabs, makecell, arydshln, rotating}
\usepackage{pifont}
\usepackage{wrapfig}
\usepackage{lipsum}
\usepackage{amsmath}
\usepackage{amssymb}
\usepackage{comment}
\usepackage{mathtools}
\usepackage{balance}
\usepackage{subcaption}
\usepackage{caption}
\usepackage{relsize}
\usepackage{multirow}
\usepackage{nccmath}
\usepackage{epsfig}
\usepackage{amsthm,mathrsfs,bm}
\usepackage{dsfont,color,bbm}
\usepackage{algorithm}
\usepackage{resizegather}
\usepackage{booktabs}
\usepackage{float}
\usepackage{xr}
\usepackage{enumitem}
\usepackage{times}
\usepackage{latexsym}
\usepackage{lipsum}
\usepackage{pdflscape}
\usepackage{booktabs}
\usepackage{multirow}
\usepackage{tcolorbox}
\usepackage{arydshln}
\tcbuselibrary{breakable}

\renewcommand{\algorithmiccomment}[1]{\bgroup\hfill\tiny//~#1\egroup}

\definecolor{cvprblue}{rgb}{0.21,0.49,0.74}
\usepackage[pagebackref,breaklinks,colorlinks,allcolors=cvprblue]{hyperref}

\def\paperID{40634} 
\def\confName{CVPR}
\def\confYear{2026}

\title{Attention-aware Inference Optimizations for Large Vision-Language Models \\ 
with Memory-efficient Decoding}

\author{Fatih Ilhan\textsuperscript{1}, Gaowen Liu\textsuperscript{2}, Ramana Rao Kompella\textsuperscript{2}, \\ Selim Furkan Tekin\textsuperscript{1} 
Tiansheng Huang\textsuperscript{1}, Zachary Yahn\textsuperscript{1}, Yichang Xu\textsuperscript{1}, Ling Liu\textsuperscript{1}\\
\textsuperscript{1}Georgia Institute of Technology, \textsuperscript{2}Cisco Research \\
{\tt\small
filhan@gatech.edu, 
\{gaoliu,rkompell\}@cisco.com,} \\
{\tt\small\{stekin6,thuang,zachary.yahn,xuyichang\}@gatech.edu,
ling.liu@cc.gatech.edu}}

\begin{document}
\maketitle
\begin{abstract}
Large Vision-Language Models (VLMs) have achieved remarkable success in multi-modal reasoning, but their inference time efficiency remains a significant challenge due to the memory overhead during decoding, especially when the query and answer of VLMs consist of long sequences of visual and text tokens. This paper presents AttentionPack, an adaptive and attention-aware optimization framework tailored for large vision-language models with improving memory-efficiency during decoding, focusing on addressing the challenges due to the increased high number of visual inputs and interactions, particularly in long-context tasks with multiple high-resolution images or videos.
AttentionPack is novel in two aspects:
(i) We introduce a multi-head attention compaction method for economically storing key and value matrices by exploiting the implicit low-rank structure, and (ii) we develop a token-specific attention-aware decompression mechanism to reduce latency overhead. 
Experimental results on multiple benchmarks demonstrate that AttentionPack improves memory efficiency by up to 8x, enabling higher batch sizes and faster batch inference while preserving the model output quality or longer context lengths for superior retrieval performance. We also report the effectiveness of AttentionPack combined with eviction, quantization and kernel fusion, showing further efficiency gains for resource-limited environments.
\vspace{-10pt}
\end{abstract}
\section{Introduction}
\label{sec:intro}

Large Vision-Language Models (VLMs) with billions of parameters trained on vast amounts of multi-modal data have achieved remarkable success across both computer vision and natural language processing domains~\cite{llava, gemini}. These models are reshaping how we interact with AI systems in ways that cannot be achieved solely by language or vision uni-modal models. Currently, a majority of popular large VLMs rely on sequential visual representation, where images are transformed into hundreds or thousands of tokens when feeding them to LLMs along with language prompts~\cite{llava, videollava}.

However, large VLM inference can be prohibitively expensive, especially for tasks requiring long context, such as in dialogue systems, document understanding and video analysis. A significant amount of temporary intermediate data (containing the key and value vectors of past tokens for each batch instance, layer and attention head) is stored to prevent recomputation during decoding, known as KV cache, with a size increasing with the sequence length, number of dimensions and batch size. Although this technique reduces computation during a single decoding step, one needs to fetch ever-growing data, only to perform vector-to-matrix operations. This may result in more time being spent loading these vectors to GPU memory rather than for the computation itself, especially for long-context tasks, causing under-utilization of the compute and high inference latency. For instance, an large VLM with 13B parameters~\cite{llava}, processing 16 images each with 256 tokens and a batch size of 64, may require around 214 GB for inference at half precision. To this end, in this work, we tackle the challenges of efficient storage, compression and retrieval of these intermediate structured data under stringent memory constraints.

Recent studies aiming to directly reduce the size of this cache can be categorized into two directions: (i) token eviction along the sequential axis, and (ii) compression through quantization to represent information with fewer bits. Eviction techniques~\cite{h2o,scissor} focus on developing algorithms to release tokens from the cache based on certain heuristic criteria such as attention scores, norm of key vectors, or token categories. However, the memory footprint reduction remains limited as the number of stored dimensions remains the same. Quantization-based approaches~\cite{kvquant,gear} represent the vectors with lower bit precision but struggles with outlier values and hardware compatibility. large VLMs leverage the advanced emergent capabilities inherent in their language components, however, only a few recent works focus on analyzing how the behavior of visual tokens differ and capitalizing these differences to enhance efficiency~\cite{fastv}. To this end, we focus on thoroughly understanding this aspect, specifically, analyzing the low-rank structure of visual/textual tokens and significantly reducing the memory footprint by compressing them along the hidden dimension axis without evicting tokens.

AttentionPack capitalizes on the fact that stored key and value vectors, particularly for visual tokens, tend to have low intrinsic rank, which can be utilized to represent the data with fewer dimensions using singular value decomposition (SVD). Based on our analysis, we conclude that these vectors can be compressed up to 8 times for large VLMs such as LLaVA variants and QwenVL without hurting model output quality. In addition, we develop an attention-aware decompression technique to reduce the overhead due to decompression at decoding stage. Thanks to the reduced cache size, users can operate with larger batch sizes to improve throughput or increase context length to enhance retrieval. We perform experiments on various image and video tasks with LLaVA1.5~\cite{llava}, QwenVL~\cite{qwen} and VideoLLaVA~\cite{videollava} variants, which demonstrates the effectiveness of our approach. We further analyze the behavior of compression and decompression, their sensitivity to different parameters, latency breakdown and compatibility with other techniques.
\section{Related Work}
\label{sec:rw}

A wide range of prior research efforts have focused on either architectural modifications or fine-tuning adjustments~\cite{reformer,gqa} for efficiency. For instance, approaches such as Reformer~\cite{reformer}, Longformer~\cite{longformer}, and Performer~\cite{performer} aim to improve the scalability of attention mechanisms when dealing with long sequences by token grouping and approximating attention patterns. In the vision domain, A-ViT~\cite{avit} proposes a learnable token dropping mechanism for vision transformers during inference. While these techniques offer reductions in the computational cost of attention operations, they do not address the memory bottlenecks that arise from storing the key-value cache. Notably, grouped-query attention (GQA)\cite{gqa} reduces the cache footprint and accelerates inference speed by sharing the same key and value matrices across heads. Other efforts have explored improving memory management and I/O efficiency to reduce latency. Low-level optimizations such as FlashAttention~\cite{flashattention} and PageAttention~\cite{pagedattention} minimizes costly memory transfers during attention computation or optimizes resource allocation using paging techniques. Nevertheless, despite these advancements, the size of the cache remains same, continuing to pose a major source of memory bottleneck for long-context inference.

In addition to these strategies, model compression techniques such as pruning~\cite{llmpruner,recap} and low-rank approximations~\cite{asvd} have been explored as indirect methods to reduce the size of the key-value cache by shrinking the hidden dimension size of the model’s weights. However, these methods generally require extensive pre-deployment optimization on a representative dataset to preserve model performance, posing limitations for users who seek to run inference without engaging in additional model fine-tuning. To further alleviate cache memory constraints, token merging~\cite{mini} or quantization-based techniques have also been proposed~\cite{kvquant,gear}. For example, KVQuant~\cite{kvquant} introduces a sensitivity-aware, non-uniform quantization strategy that operates after outlier values in the cache have been eliminated. Similarly, GEAR~\cite{gear} leverages a SVD-based decomposition of residuals, enabling more accurate reconstruction of dequantized key-value vectors. While effective, these approaches often rely on hardware-specific low-level optimizations, and assumptions regarding the minimal impact of outliers or newly generated vectors on overall performance.

\begin{figure*}[t!]
    \centering
    \includegraphics[width=0.88\textwidth]{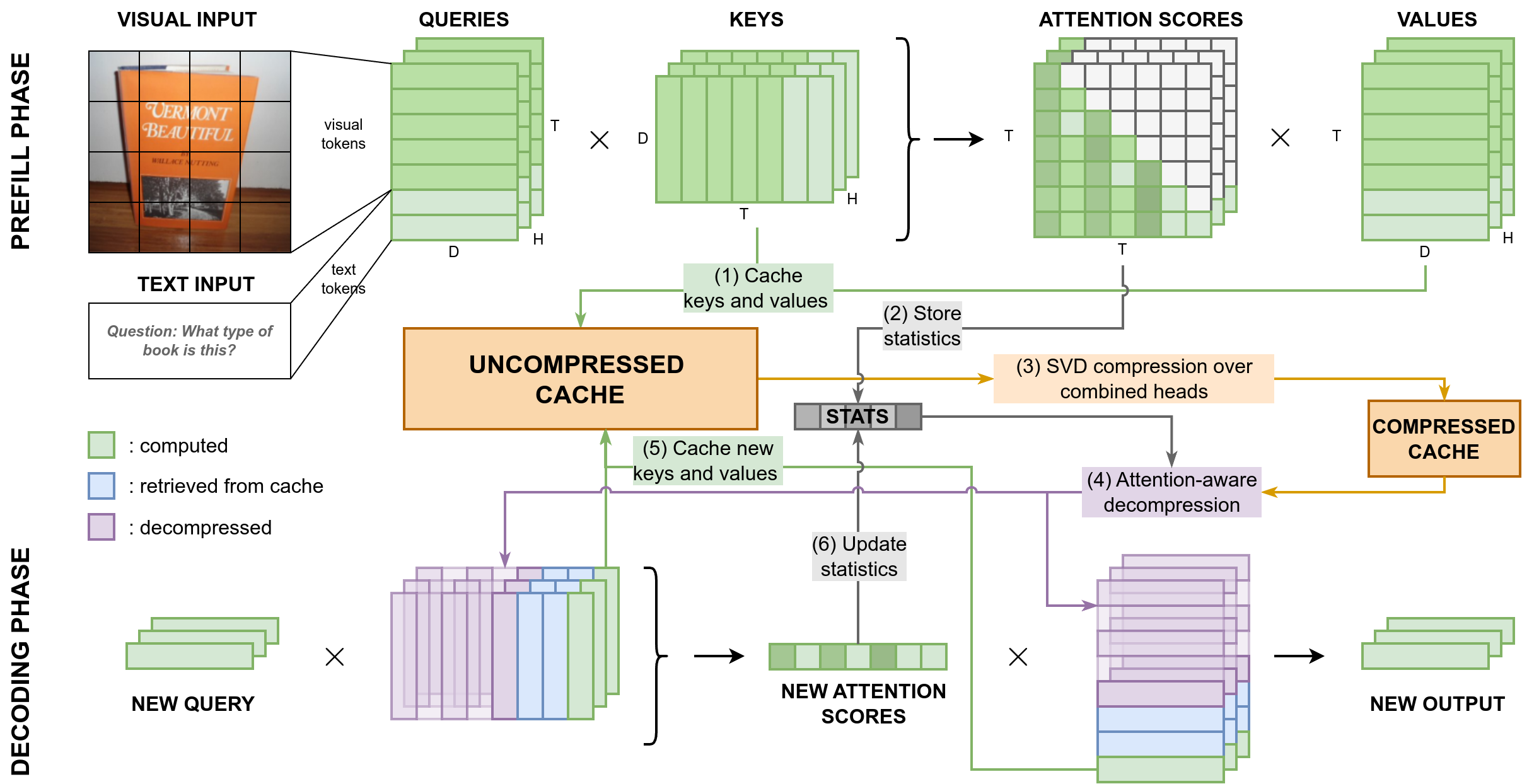}
    \caption{The schematic of the workflow during inference. After the prefill phase, at each decoding step, we first compress the cache along combined heads. We perform attention-aware partial decompression before attention score computation.}
    \label{fig:main}
    \vspace{-8pt}
\end{figure*}

In recent years, research efforts has focused toward token eviction policies designed to manage the memory for key-value cache more effectively by selectively discarding certain tokens. H2O~\cite{h2o}, for example, implements a sliding window approach that retains only the most recent tokens while discarding others based on accumulated attention scores. Scissorhands~\cite{scissor} follows a different strategy by identifying and removing tokens that consistently receive low attention scores during inference. In the specific case of LVLMs, several studies have investigated redundancy within visual tokens and proposed token dropping techniques tailored to this modality. FastV~\cite{fastv} introduces an eviction criterion based on attention scores computed in the final decoding step, while Look-M~\cite{lookm} applies eviction based on cumulative attention scores accumulated over time. Despite their promising results, these eviction-based methods face challenges in substantially reducing memory usage due to key-value cache without introducing unacceptable levels of information loss, as the hidden dimension size of cached vectors remains constant. To overcome this limitation, recent work has explored decomposition of key and value weight matrices to reduce cache size along the hidden dimensions~\cite{palu}. 
\section{Methodology}
\label{sec:method}

This section provides a comprehensive overview of our methodology. We begin with a formal description of the key-value caching process in LVLMs, followed by an analysis on the low-rank structure of stored vectors. Next, we detail the multi-head compression of these vectors using matrix decomposition. Lastly, we thoroughly explain the attention-aware decompression stage and provide an algorithmic description of the overall process during decoding.  We provide a detailed algorithmic description in Appendix~\ref{sec:algo}.

Consider an LVLM such as LLaVA~\cite{llava} which accepts visual $\mathbf{X_v}$ and text inputs $\mathbf{X_t}$. A pretrained visual encoder can be used as backbone to extract features from visual inputs $\mathbf{Z_v}=g(\mathbf{X_v}) \in \mathbb{R} \in {T_v \times D_v}$, followed by a linear projection using matrix $\mathbf{W}_p \in \mathbb{R}^{D_v \times HD}$, $\mathbf{H_v}=\mathbf{W}_p\mathbf{X_v}$ to achieve the same dimensionality with embedded text input vectors $\mathbf{H_t} \in \mathbb{R}^{T_t \times HD}$. Here, $T_t$ and $T_v$ are the number of textual and visual tokens. In the attention block within a layer, weights \(\boldsymbol{\theta} = \{\mathbf{W}_q, \mathbf{W}_k, \mathbf{W}_v, \mathbf{W}_o \in \mathbb{R}^{HD \times HD}\}\) are used for query, key, value, and output transformations. Here, \(H\) and \(D\) are the number of heads and hidden dimensions.

During decoding, attention computations require access to key and value vectors from past. These vectors are stored in a key-value cache, represented as \(\mathcal{C} = \{\mathbf{K}, \mathbf{V} \in \mathbb{R}^{T \times HD}\}\), where \(T=T_v+T_t\) is the number of stored tokens after the prefill phase. At each decoding step, new key \(\boldsymbol{k}\) and value \(\boldsymbol{v}\) vectors are appended to the cache and used for attention calculations. While this technique reduces computational overhead, it potentially leads to memory constraints in long-context scenarios as storing these vectors in system memory introduces significant latency due to data transfers between host (CPU) and device (GPU) memory. To address this, our approach compresses these vectors, enabling efficient storage in GPU memory and eliminating offloading overhead. Figure~\ref{fig:main} provides an overview of the workflow.

\subsection{Multi-head Compression of Key/Value Vectors}
\begin{figure}[t!]
    \includegraphics[width=\columnwidth]{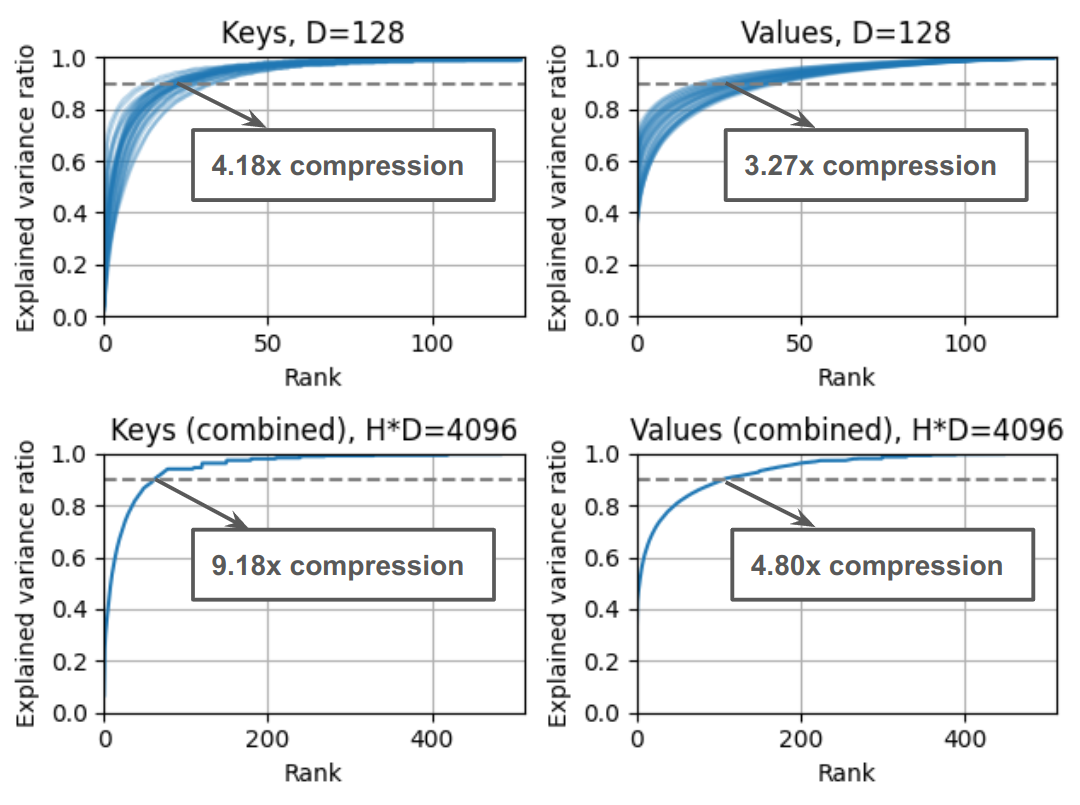}
    \vspace{-10pt}
    \caption{Rank vs explained variance ratio without/with combining along head axis before compression for key and value vectors.}
    \label{fig:lr}
    \vspace{-10pt}
\end{figure}

Reducing the size of key-value cache allows for (i) processing queries in larger batches, leveraging parallelization to minimize overall processing time, and (ii) enabling long-context inference in GPU-constrained environments. To achieve this, we apply compression over the uncompressed key and value vectors to significantly shrink their sizes. Our analysis on the first layer of LLaVA1.5-7B~\cite{llava} over OCR-VQA~\cite{ocrvqa} samples ($T_v=576$), as illustrated in Figure~\ref{fig:lr}, demonstrates that the stored vectors exhibit an inherent low-rank structure, which can be exploited for dimensionality reduction using singular value decomposition (SVD). Additionally, we find that compression is more effective when low-rank decomposition is applied to the combined cache along multiple heads rather than treating each head independently. However, handling visual and textual tokens separately is critical, as they originate from different modalities and applying SVD indiscriminately across both results in suboptimal compression. To this end, we apply the procedures separately for visual/textual input tokens. To achieve higher compression rates, before applying low-rank SVD, we first merge the vectors along the head axis, allowing shared information across heads to be compressed together.

\begin{figure}[t!]    \includegraphics[width=.95\columnwidth]{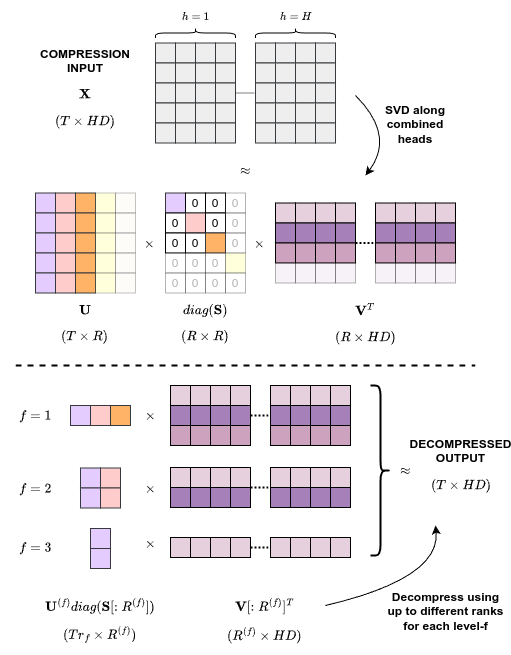}
    \caption{Visualization of compression and partial decompression.}
    \label{fig:svd}
    \vspace{-10pt}
\end{figure}

This process decomposes each stored matrix into two low-rank components: \( \mathbf{K}_{*} \approx \overline{\mathbf{K}}_{*}\mathbf{D}_{k*} \) and \( \mathbf{V}_{*} \approx \overline{\mathbf{V}}_{*}\mathbf{D}_{v*} \), where \( \overline{\mathbf{K}}_{*} \in \mathbb{R}^{T_{*}\times R_{k*}} \) represents the compressed key cache, \( \mathbf{D}_{k*} \in \mathbb{R}^{R_{k*} \times HD} \) serves as the key decompression matrix, \( \overline{\mathbf{V}}_{*} \in \mathbb{R}^{T_{*}\times R_{v*}} \) is the compressed value cache, and \( \mathbf{D}_{v*} \in \mathbb{R}^{R_{v*} \times HD} \) serves as the value decompression matrix. Each column of the first low-rank component is scaled using the computed singular values after SVD. Here to notate, ${*}=v$ for visual tokens and ${*}=t$ for textual tokens. The rank values \( R_{kv}, R_{kt} \) and \( R_{vv}, R_{vt} \) for key and value matrices are tunable parameters and allow adjustments based on the available memory budget. We illustrate the compression process in the top half of Figure~\ref{fig:svd} (Number of dimensions in the figure are for illustration purposes). For instance, the storage requirement for key vectors of visual tokens at a single layer and for one input decreases from \( T_vHD \) to \( T_vR_{kv} + R_{kv}HD\) after applying compression for $R_{kv} << \min(T_v, HD)$, resulting in the following compression ratio: $
    c_{kv} = \frac{T_vHD}{T_vR_{kv} + R_{kv}HD} $.
As a numerical example, when \( T_v=1000 \), \( H=40 \), \( D=128 \), and \( R_{kv} = 64 \), the resulting memory footprint after compression is around 13x lower.

\subsection{Attention-aware Decompression}
Before computing attention scores, the compressed vectors are decompressed by multiplying them with the corresponding decompression matrices. While compression is performed once after the prefill phase and periodically during decoding, decompression occurs at every step and introduces latency overhead. However, the memory savings from compression enable larger batches, allowing parallelization to mitigate this overhead and speed up overall computation. In single-instance inference, however, the added latency can increase by up to 30\%, depending on the context length.

To mitigate this overhead, we introduce attention-aware decompression, leveraging that not all stored tokens contribute equally at every decoding step. While all tokens are initially compressed to the same low rank, fewer ranks can be used during decompression for tokens (image regions in visual input and words in textual input) that are less influential in the final output computation. To implement this, we track scaled accumulated attention scores using moving average with parameter $\alpha \in [0, 1)$ after computing attention scores at each step. At decoding step \( t \), the importance score for each token at step \( t_p \) for $t_p \in \{1, ..., t\}$ is computed as: \noindent \vspace{-5pt}
\begin{equation}
    \mathcal{I}_{t_p} \leftarrow \alpha^{T_q}\mathcal{I}_{t_p} + (1-\alpha^{T_q})\frac{\sum_{t'=t-T_q}^t A_{t't_p}}{T_q},
    \label{eq:imp} \vspace{-10pt}
\end{equation} \noindent
where $T_q$ is the number of new tokens, which is one except an external input with multiple tokens is provided by the user. Here, $A_{t't_p}$ represents the attention weight from $t_p$ in decoding step $t'$, and averaged across heads. Setting $\alpha=0$ only takes the most recent attention score into account.

Tokens with the highest scores are decompressed using the original compression rank, while those with lower scores are decompressed with reduced ranks. This strategy significantly decreases the decompression FLOPs. For instance, in the key cache for visual tokens, this reduces the FLOPs from \( 2 T_v HD R_{kv} \) to \( 2 T_v HD\sum_{f=1}^F r_f R_{kv}^{(f)} \), where $F$ is the number of groups, each with $r_fT_v$ tokens and decompression rank of $R_{kv}^{(f)}$. We set the group $f=1$ having the highest importance and decompress with $R_{kv}^{(1)}=R_{kv}$, i.e. the rank used in compression. As an example, for $F=2$, $T=1000$, $r_1=0.1$, $r_2=0.9$, $R_{kv}^{(1)}=64$, $R_{kv}^{(2)}=16$, where the 90\% of tokens with the lowest scores are decompressed with the 25\% of the rank, the FLOPs reduce by 67.5\%.
\vspace{-2pt}
\begin{table*}[t!]
\setlength{\tabcolsep}{5pt}
\centering
\begin{adjustbox}{width=\textwidth}
\begin{tabular}{@{}clccccc@{}}
\toprule
\multirow{3}{*}{\textbf{Model}} & \multirow{3}{*}{\textbf{Method}} & \textbf{Cache Memory} & \textbf{Average} & \multicolumn{3}{c}{\textbf{Image QA Datasets}} \\
 & & \textbf{Reduction} & \textbf{Throughput} & A-OKVQA & OCR-VQA & MMMU \\
 & & (avg. per instance) & \textbf{Change} & (acc.) & (rouge-l) & (acc.) \\ \midrule
\multirow{7}{*}{LLaVA1.5-7B} & Full KV Caching & $-$ & $-$ & $76.64$ & $51.05$ & $34.68$ \\
 \cdashline{2-7}\noalign{\vskip 0.4ex}
 & FastV~\cite{fastv} ($k=5$, $e=50\%$) & $1\times$ & $+29\%$ & $\underline{76.72}$ & $\bm{52.62}$ & $\underline{34.27}$\\
 & ScissorHands~\cite{scissor} ($e=50\%$) &  $2\times$ & $+32\%$ & $75.50$ & $50.08$ & $31.66$\\
 & H2O~\cite{h2o} ($e=50\%$) &  $2\times$ & $+32\%$ & $76.36$ & $52.09$ & $32.15$ \\
 & Minicache~\cite{mini}  & $4.51\times$ & $+44\%$ & $76.54$ & $51.93$ & $33.75$ \\
 & \textbf{AttentionPack} ($R_{kv}=R_{vv}=64$) & $\underline{5.09\times}$ & $\underline{+54\%}$ & $\bm{76.88}$ & $\underline{52.44}$ & $\bm{34.59}$\\
 & \textbf{AttentionPack} ($R_{kv}=R_{vv}=32$) & $\bm{7.24\times}$ & $\bm{+65\%}$ & ${75.91}$ & ${51.69}$ & ${32.79}$\\
 \midrule
\multirow{7}{*}{LLaVA1.5-13B} & Full KV Caching & $-$& $-$ & $81.40$ & $55.01$ & $36.40$\\
 \cdashline{2-7}\noalign{\vskip 0.4ex}
 & FastV~\cite{fastv} ($k=5$, $e=50\%$) & $1\times$ & $+20\%$ & ${81.17}$ & $\bm{53.50}$ & $\bm{36.61}$\\
 & ScissorHands~\cite{scissor} ($e=50\%$) & $2\times$ & $+24\%$ & $77.48$ & $51.32$ & $31.90$\\
 & H2O~\cite{h2o} ($e=50\%$) & $2\times$ & $+24\%$ & $79.10$ & $52.49$ & $33.01$\\
 & Minicache~\cite{mini} & $4.35\times$ & $+38\%$ & $\underline{81.20}$ & $53.18$ & $35.52$ \\
 & \textbf{AttentionPack} ($R_{kv}=R_{vv}=64$) & $\underline{5.17\times}$ & $\underline{+43\%}$ & $\bm{81.25}$ & $\underline{53.22}$ & $\underline{36.38}$\\
 & \textbf{AttentionPack} ($R_{kv}=R_{vv}=32$) & $\bm{7.30\times}$ & $\bm{+50\%}$ & ${80.00}$ & ${51.92}$ & ${34.40}$\\
 \midrule
\multirow{6}{*}{QwenVL-Chat-7B} & Full KV Caching & $-$& $-$ & $75.67$ & $70.5$ & $35.82$\\
 \cdashline{2-7}\noalign{\vskip 0.4ex}
 & FastV~\cite{fastv} ($k=2$, $e=50\%$) & $1\times$ & $+45\%$ & $75.10$ & $66.43$ & $\bm{35.80}$\\
 & ScissorHands~\cite{scissor} ($e=50\%$) & $2\times$ & $+51\%$ & $74.60$ & $67.10$ & $35.14$\\
 & H2O~\cite{h2o} ($e=50\%$) & $2\times$ & $+51\%$ & $74.98$ & $67.15$ & $35.05$\\
 & \textbf{AttentionPack} ($R_{kv}=R_{vv}=64$) & $\underline{2.77\times}$ & $\underline{+61\%}$ & $\bm{75.33}$ & $\bm{68.45}$ & $\underline{35.72}$\\
 & \textbf{AttentionPack} ($R_{kv}=R_{vv}=32$) & $\bm{4.02\times}$ & $\bm{+74\%}$ & $\underline{75.17}$ & $\underline{68.02}$ & $35.67$\\
 \bottomrule
\end{tabular}
\end{adjustbox}
\caption{Image QA results on AOKV-QA, OCRKV-QA and MMMU with LLaVA1.5-7B, LLaVA1.5-13B and QwenVL-Chat-7B. We report evaluation metrics along with cache memory and throughput statistics for full KV caching, ScissorHands~\cite{scissor}, H2O~\cite{h2o}, FastV~\cite{fastv}, Minicache~\cite{mini} to compare with AttentionPack.}
\vspace{-10pt}
\label{tab:main_image}
\end{table*}

\section{Results}
\label{sec:results}

In this section, we present the results obtained across various datasets, along with comparisons to state-of-the-art efficiency techniques and an in-depth analysis of memory and computational costs. To summarize, our approach achieves around 80\% reduction in cache size (5.1x smaller) on LLaVA1.5-7B and LLaVA1.5-13B, 63\% reduction (2.8x smaller) on QwenVL-Chat-7B and 88\% reduction (8.1x smaller) on VideoLLaVA through compression while maintaining performance. The reduced memory footprint enables up to 74\% faster decoding in batch inference with attention-aware decompression for image and 60\% for video QA. 

\subsection{Experiment Setup}

We experiment on five datasets from different image/video tasks A-OKVQA~\cite{aokvqa}, commonsense and world knowledge task, (ii) OCR-VQA~\cite{ocrvqa}, optical character recognition task, (iii) MMMU~\cite{mmmu} a multi-discipline multi-modal
reasoning benchmark, (iv) MSVD-QA~\cite{msvdqa} and (v) MSRVTT-QA~\cite{msvdqa} containing short video clips with multiple question-answer pairs for each video. For image QA tasks, we experiment with 7B and 13B variants of the LLaVA1.5~\cite{llava} and QwenVL-Chat-7B~\cite{qwen}. We also consider Qwen3VL-8B-instruct for experiments with grouped query attention~\cite{qwen3vl}. In video QA tasks, we consider VideoLLaVA-7B, which has a similar architecture with LLaVA. For comparisons, we consider: H2O~\cite{h2o}, Scissorhands~\cite{scissor}, FastV~\cite{fastv} and Minicache~\cite{mini}. We perform attention-aware decompression over value cache with $F=2, r_1=0.25, r_2=0.75, R_{vv}=R_{vv}^{(1)}=4R_{vv}^{(1)}$. In initial experiments, $\alpha \in [0.05, 0.75]$ gave robust and stable results and we set $\alpha=0.25$. 

\begin{table*}[t!]
\setlength{\tabcolsep}{5pt}
\centering
\begin{adjustbox}{width=\textwidth}
\begin{tabular}{@{}clccccc@{}}
\toprule
\multirow{3}{*}{\textbf{Model}} & \multirow{3}{*}{\textbf{Method}} & \multicolumn{2}{c}{\begin{tabular}[c]{@{}c@{}}\textbf{Cache Memory} \end{tabular}} & \textbf{Batch Inf.} & \multicolumn{2}{c}{\textbf{Video QA Datasets}} \\
 &  & \multicolumn{2}{c}{\begin{tabular}[c]{@{}c@{}} (avg. per instance) \end{tabular}} & \textbf{Average}  & MSVD-QA & MSRVTT-QA \\
 & & Size (MB) & Reduction & \textbf{Change} & (acc.) & (acc.) \\ \midrule
\multirow{5}{*}{VideoLLaVA-7B} & Full KV Caching & $1114.6$ & $1\times$ & $-$ & $69.33$ & $55.60$\\
 \cdashline{2-7}\noalign{\vskip 0.4ex}
 & FastV~\cite{fastv} ($k=5$, $e=50\%$) & $1114.6$ & $1\times$ & $+23\%$ & $\bm{69.60}$ & $\bm{55.52}$\\
 & Scissorhands~\cite{scissor} ($e=50\%$) & $557.3$ & $2\times$ & $+32\%$ & $65.90$ & $54.33$\\
 & H2O~\cite{h2o} ($e=50\%$) & $557.3$ & $2\times$ & $+32\%$ & $66.54$ & $54.80$\\
 & \textbf{AttentionPack} ($R_{kv}=R_{vv}=128$) & $\bm{137.5}$ & $\bm{8.11\times}$ & $\bm{+60\%}$ & $\underline{69.21}$ & $\underline{55.47}$ \\
\bottomrule
\end{tabular}
\end{adjustbox}
\caption{Video QA results on MSVD-QA and MSRVTT-QA with VideoLLaVA-7B. We report evaluation metrics along with cache memory and throughput statistics for full KV caching, H2O eviction, FastV to compare with our approach AttentionPack.}
\label{tab:main_video}
\end{table*}

\subsection{Comparisons}
\label{sec:exp_main}
We report the main experiment results in  Tables~\ref{tab:main_image} and and~\ref{tab:main_video} for image and video question-answering (QA) tasks.
For QA tasks with text output we report ROUGE-\textit{L} (F1) scores, where \textit{L} is the length of the longest common subsequence between prediction and target. For multiple choice question tasks, we report accuracy. We also report average cache size per instance and decoding throughput change compared to baseline with batch inference. For QA pair instances corresponding to the same image/video, we perform inference together without releasing the key-value cache. For H2O and ScissorHands, we evict 50\% of tokens following the algorithms on attention-score based eviction policies. In Minicache, we follow the settings applied for LLaMA variants from the paper~\cite{mini}, on the visual tokens. In FastV, we apply 50\% token skipping starting after the second or fifth layer but as the tokens are not released and memory usage due to the key-value cache stays the same.

With LLaVA1.5-7B, AttentionPack performs +0.16\%, -0.18\% and +0.32\% with respect to the closest comparison across three datasets while enabling 5.09 times smaller cache size on average per instance. For instance, the total (textual+visual) size for full KV caching per sample is around 328.2 MB, and is reduced to 64.5 MB with AttentionPack. Likewise, with LLaVA1.5-13B, AttentionPack performs +0.08\%, -0.28\% and -0.23\% with respect to the closest comparison across three datasets while enabling 5.17 times smaller cache size. Lastly, with QwenVL-Chat-7B, AttentionPack performs +0.35\%, +1.30\% and -0.08\% across three datasets while enabling 2.77 times smaller cache size. Reduced memory footprint with AttentionPack enables larger batch sizes, increasing the throughput in batch inference up to 61\% for the rank of 64. We have similar observation for video QA as reported in Table~\ref{tab:main_video}. AttentionPack performs 0.39\% and 0.05\% worse compared to FastV while having 8.11 times smaller cache. In video QA, the size of visual token cache increases by the number of frames. Supported by the fact that most frames carry similar information, we observe that higher rates of compression is achievable without resulting in performance decrease.

\subsection{Analysis of Rank and Decompression}


\begin{table}[t!]
    \centering
    \setlength{\tabcolsep}{2pt}
    \begin{adjustbox}{width=\columnwidth}
    \begin{tabular}{@{}cccccccc@{}}
    \toprule
    \multicolumn{2}{c}{\textbf{Key Cache}} & \multicolumn{2}{c}{\textbf{Value Cache}} & \multicolumn{2}{c}{\textbf{Total Cache}} & \multicolumn{2}{c}{\textbf{Dataset}} \\
    $R_{kv}$ & Red. & $R_{vv}$ & Red. & Size (GB) & Red. & \small{AOKVQA} & \small{OCRVQA} \\ \midrule
    & $-$ & $-$ & $-$ & $10.50$ & $-$ & $76.64$ & $51.05$\\ 
     \cdashline{1-8}\noalign{\vskip 0.4ex}
    \multirow{2}{*}{$128$} & \multirow{2}{*}{$3.19\times$} & $256$ & $1.83\times$ & $4.51$ & $2.32\times$ & $76.77$ & $52.50$\\
     &  & $128$ & $3.19\times$ & $3.28$ & $3.19\times$ & $76.80$ & $52.51$\\ 
     \cdashline{1-8}\noalign{\vskip 0.4ex}
    \multirow{2}{*}{$64$} & \multirow{2}{*}{$5.09\times$} & $128$ &$3.19\times$ & $2.68$ & $3.92\times$ & $76.75$ & $52.47$ \\
     &  & $64$ & $5.09\times$ & $2.06$ & $5.09\times$ & $76.88$ & $52.44$ \\
    \cdashline{1-8}\noalign{\vskip 0.4ex}
    \multirow{2}{*}{$32$} & \multirow{2}{*}{$7.24\times$} & $64$ & $5.09\times$ & $1.75$ & $5.98\times$ & $76.46$ & $52.00$ \\
     &  & $32$ & $7.24\times$ & $1.45$ & $7.24\times$ & $75.91$ & $51.69$ \\ 
     \cdashline{1-8}\noalign{\vskip 0.4ex}
     \multirow{2}{*}{$16$} & \multirow{2}{*}{$9.18\times$} & $32$ & $7.24\times$ & $1.29$ & $8.09\times$ & $74.12$ & $50.06$ \\
     &  & $16$ & $9.18\times$ & $1.14$ & $9.18\times$ & $72.13$ & $48.44$ \\ 
     \cdashline{1-8}\noalign{\vskip 0.4ex}
     \multirow{2}{*}{$16\rightarrow128$} & \multirow{2}{*}{$4.78\times$} & $32\rightarrow256$ & $2.95\times$ & $2.87$ & $3.65\times$ & $74.33$ & $50.85$ \\
     &  & $16\rightarrow128$ & $4.78\times$ & $2.19$ & $4.78\times$ & $72.51$ & $48.80$ \\ 
     \cdashline{1-8}\noalign{\vskip 0.4ex}
     $64$ $(90\%)$ & $5.31\times$ & $64$ $(90\%)$ & $5.16\times$ & $2.00$ & $5.23\times$ & $76.50$ & $52.18$ \\
     \bottomrule
    \end{tabular}
    \end{adjustbox}
    \caption{Performance and cache size for various compression rank values using LLaVA1.5-7B for batch size of 32.}
    \vspace{-15pt}
    \label{tab:comp}
\end{table}

\noindent
\textbf{Impact of Compression Rate: } We analyze the impact of compression rank values for key and value caches in Table~\ref{tab:comp} with LLaVA1.5-7B. We observe that for rank values below 64, the performance starts to drop for both datasets while increasing the rank from 64 to 128 brings negligible performance improvements. At $R_{kv}=R_{vv}=64$, we observe 0.24\% performance drop in A-OKVQA and 1.39\% performance increase in OCR-VQA while having around five times smaller cache, showing that based on the dataset, compression can actually filter out the irrelevant information within the visual input and enhance performance.

In addition, we reported the results for two other rank selection schemes, at the last two rows of Table~\ref{tab:comp}. First, we set the rank with a linearly increasing value such that the key cache in the first layer is compressed with the rank of 16 and last with 128. We observe that the performance is 0.38\% and 0.36\% better across two datasets compared to directly setting the rank of each layer to 16, while requiring almost twice the size of memory. We see that the ranks selected for earlier layers tend to dominate how much the model output quality will be preserved. Second, we set the rank of each layer based on the explained variance ratio such that the preserved ratio will be above 90\% if possible without exceeding a maximum rank value to 64. Based on the data, this approach may enable setting a lower rank value to certain layers with key and value vectors having intrinsic lower ranks. However, we observe around 0.3\% performance drop.

\begin{figure}[t!]
    \includegraphics[width=\columnwidth]{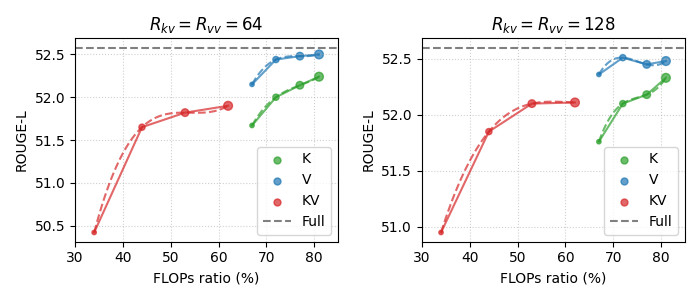}
        \caption{Impact of attention-aware decompression. Each line represents the results when AttentionPack is applied for key (k), value (v) caches or both (kv). Every line has four dots with the size of each representing the ratio of visual tokens ($r_1 \in \{0.125, 0.25, 0.375, 0.5\}$) decompressed with the full rank.}
        \label{fig:decomp}
    \vspace{-15pt}
\end{figure}

\noindent
\textbf{Impact of Attention-Aware Decompression: } We analyze the impact of attention-aware decompression in Table~\ref{fig:decomp} with LLaVA1.5-7B on OCR-VQA. Here, x-axis show the ratio of decompression FLOPs compared to the FLOPs without attention-aware decompression. Every line has four dots with the size of each representing the ratio of visual tokens ($r_1$) decompressed with the full compression rank $R_{kv}=R_{vv}=64$ (left), $R_{kv}=R_{vv}=128$ (right), while the rest ($r_2=1-r_1$) being decompressed with $R_{kv}/4$ and $R_{vv}/4$. We experiment with $r_1 \in \{0.125, 0.25, 0.375, 0.5\}$. For instance, for the compression rank of 64, we observe that decompressing the 25\% of value cache with the rank of 64 achieves almost the same performance with full decompression while having almost 30\% less FLOPs computed for decompression. However, the key vectors are more sensitive to the information loss after partial decompression as we observe higher performance drop in those scenarios, potentially due to the impact on attention weight calculations, which also affects the tokens that are fully decompressed.

\subsection{Computational Cost Analysis}
\begin{figure}[t!]
    \includegraphics[width=\columnwidth]{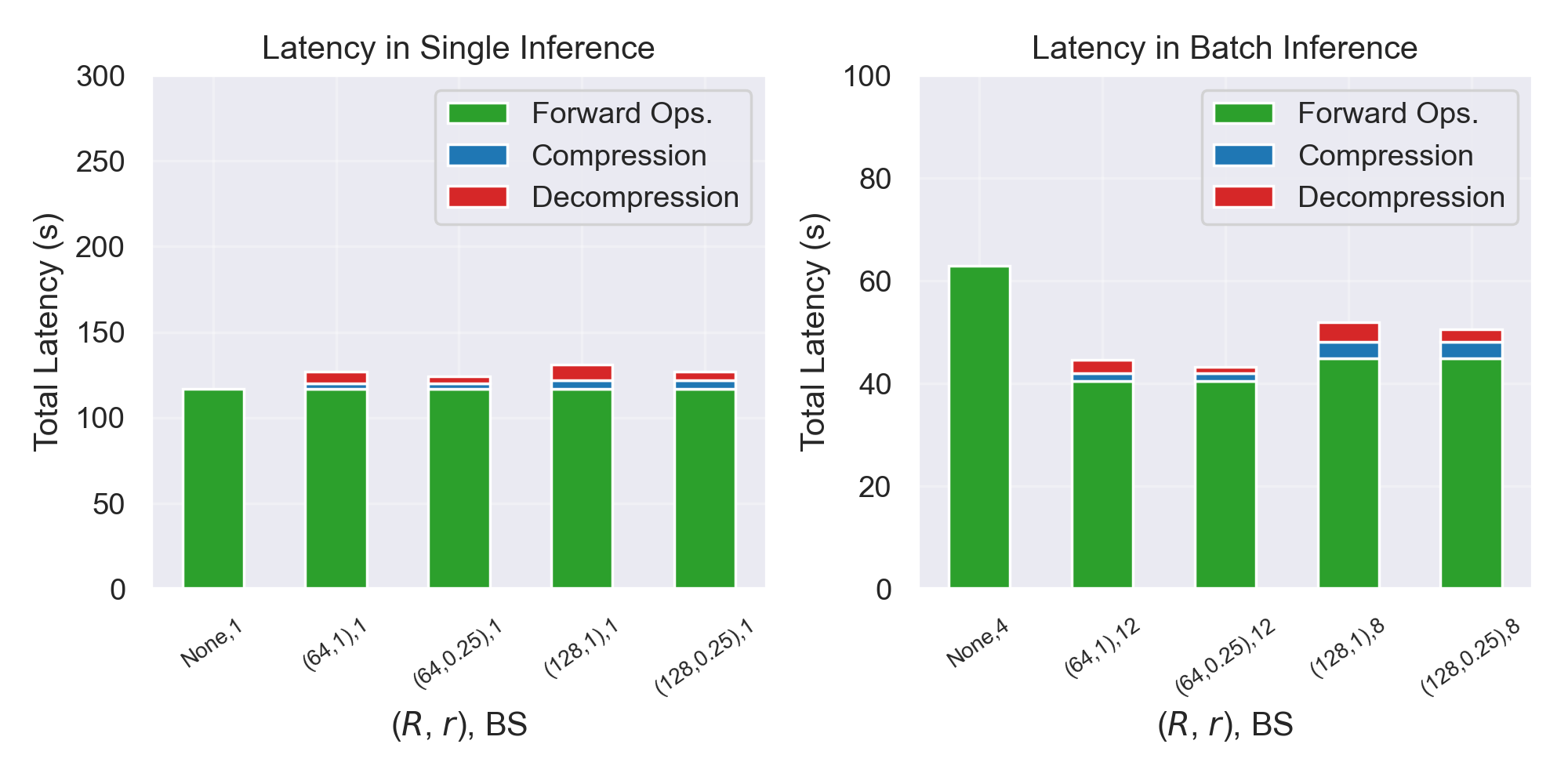}
    \vspace{-15pt}
    \caption{Total decode latency for 100 queries with LLaVA1.5-7B.}
    \vspace{-7pt}
    \label{fig:latency}
\end{figure}

In Figure~\ref{fig:latency}, we present the total decoding latency for 100 queries at OCR-VQA using LLaVA1.5-7B. We conduct measurements across different compression ranks and decompression ratios to analyze how latency varies with these configurations. The top-row plots illustrate latency in the single-inference scenario, while the bottom-row plots depict latency when the batch size is maximized to fully utilize the available GPU memory. By reducing the cache size per instance by approximately 80\% through compression, we achieve around 4 times larger batch sizes, leading to a potential reduction in total latency of up to 54\%. Latency measurements in this subsection are performed with 4-bit model weight loading and half-precision data. For these measurements, we used RTX3060, as it enabled granular monitoring of each operation (forward ops., compression, decompression) with local access. We also provide a similar analysis at various context lengths in Appendix~\ref{sec:app_latency}.

\begin{table}[t!]
\setlength{\tabcolsep}{2pt}
\centering
\begin{adjustbox}{width=\columnwidth}
\begin{tabular}{@{}lcccc@{}}
\toprule
\multirow{2}{*}{\textbf{Method}} & \multirow{2}{*}{\begin{tabular}[c]{@{}c@{}}\textbf{Avg. Cache}\\ \textbf{Size (MB)} \end{tabular}} & \multirow{2}{*}{\begin{tabular}[c]{@{}c@{}}\textbf{Avg. Thr.}\\ \textbf{Change} \end{tabular}} & \multicolumn{2}{c}{\textbf{Dataset}} \\
 &  & & A-OKVQA & OCR-VQA \\ \midrule
Full KV - fp16 & $328.2$ & $-$ & $76.64$ & $51.05$ \\
 KVQuant - 4bit~\cite{kvquant} & $82.1$ & $+49\%$ & $\underline{75.90}$ & $50.67$\\
 AttentionPack & $64.5$ & $+54\%$ & $\bm{76.88}$ & $\bm{52.44}$\\
 AttentionPack (E) & $62.1$ & $+70\%$ & $\bm{76.88}$ & $\underline{51.63}$ \\
 AttentionPack - 4bit & $\underline{16.1}$ & $\underline{+97\%}$ & $75.27$ & $50.18$ \\
 AttentionPack - 4bit (E) & $\bm{15.5}$ & $\bm{+115\%}$ & $75.27$ & $49.11$ \\
 \bottomrule
\end{tabular}
\end{adjustbox}
\caption{Collected results (accuracy on A-OKVQA rouge-l on OCR-VQA) for LLaVA1.5-7B with measured average cache size per instance (MB) and avg. throughput change in batch inference.}
\vspace{-15pt}
\label{tab:quant}
\end{table}
\begin{table}[t!]
\begin{adjustbox}{width=\columnwidth}
\centering
\begin{tabular}{@{}lcc@{}}
\toprule
\textbf{Method} & \textbf{Cache Red.} & \textbf{MMMU Acc.} \\ \midrule
Full KV Caching & - & $69.6\%$ \\
\cdashline{1-3}\noalign{\vskip 0.4ex}
FastV~\cite{fastv} ($k=2$, $e=50\%$) &  $1\times$ & $68.1\%$\\
H2O~\cite{h2o} ($e=50\%$) & $2\times$ & $66.9\%$\\
\textbf{AttentionPack} ($R_{kv}=R_{vv}=64$) & $3.17\times$ & $67.9\%$\\ \bottomrule
\end{tabular}
\end{adjustbox}
\caption{Image QA results (average cache size reduction per query vs accuracy) on MMMU with Qwen3VL-8B-instruct.}
\vspace{-7pt}
\label{tab:qwen3}
\end{table}

\begin{figure}[t!]
\centering
    \includegraphics[width=\columnwidth]{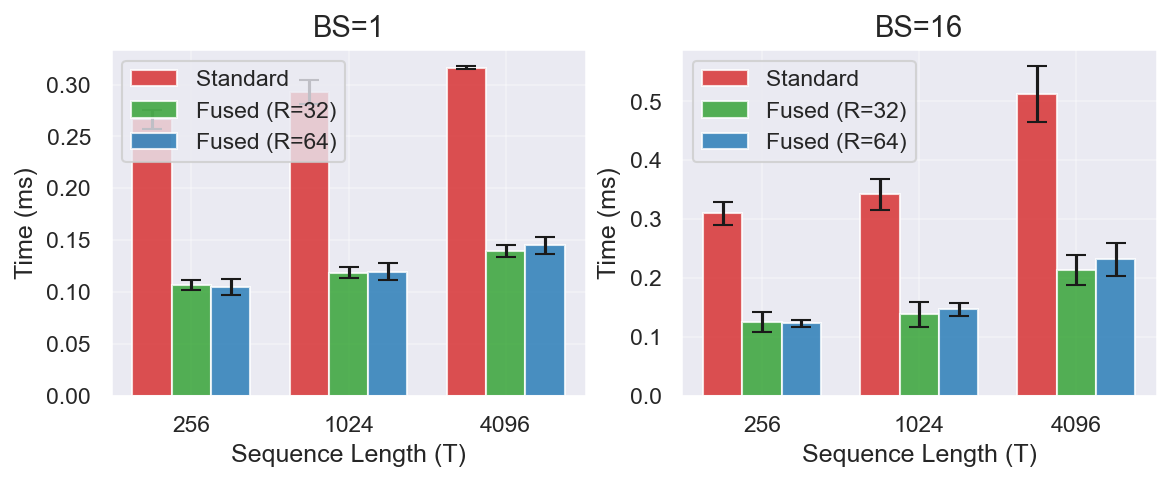}
    \caption{Decoding latency for 32 tokens with various batch sizes and sequence lengths using standard attention and with fused kernel implementation of AttentionPack. 
    We observe up to 2.4x faster inference in single and batch query settings.
    }
    \vspace{-10pt}
    \label{fig:kernel_latency}
\end{figure}

\begin{figure*}[t!]
\centering
        \subfloat{
            \includegraphics[width=\textwidth]{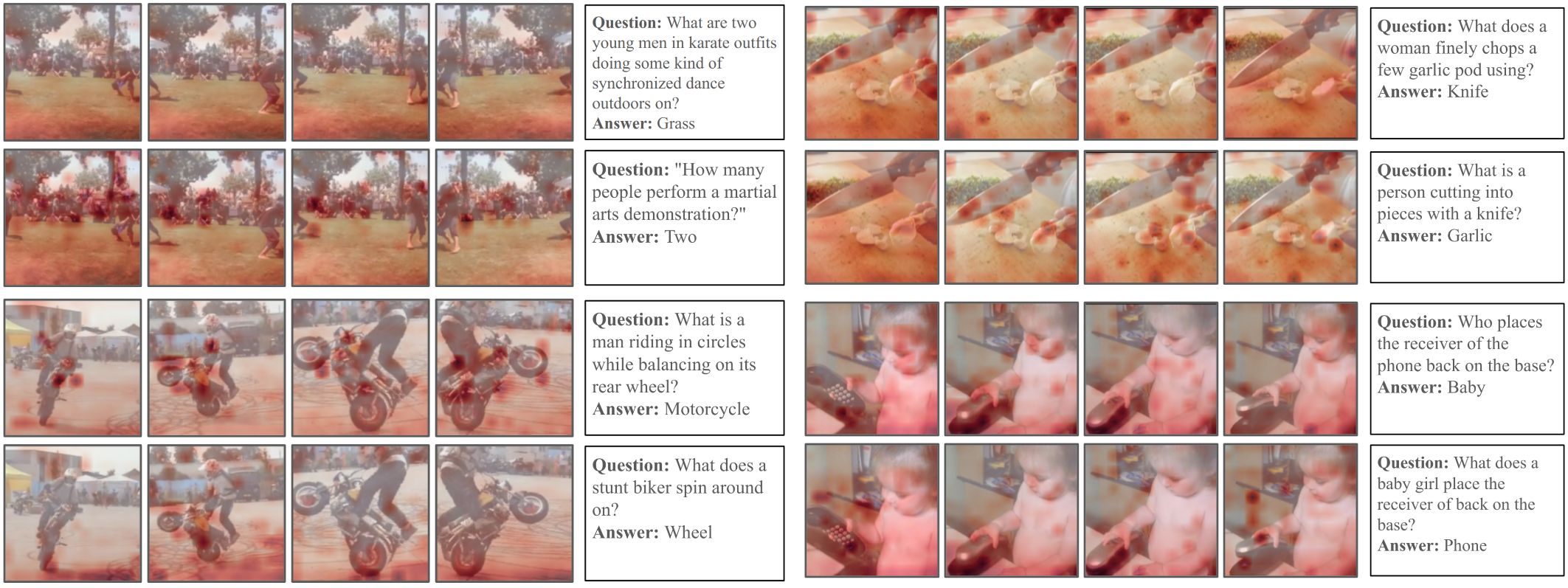}
        }
        \caption{Visualization of scores (Eq.~\ref{eq:imp}) at the first decoding step after processing the question on various examples from MSVD-QA, showing that the visual tokens closely related to the prompt have higher importance scores and will be decompressed with higher ranks.}
        \vspace{-10pt}
        \label{fig:examples_video}
\end{figure*}

\begin{figure}[t!]
\centering
        \subfloat{
            \includegraphics[width=\columnwidth]{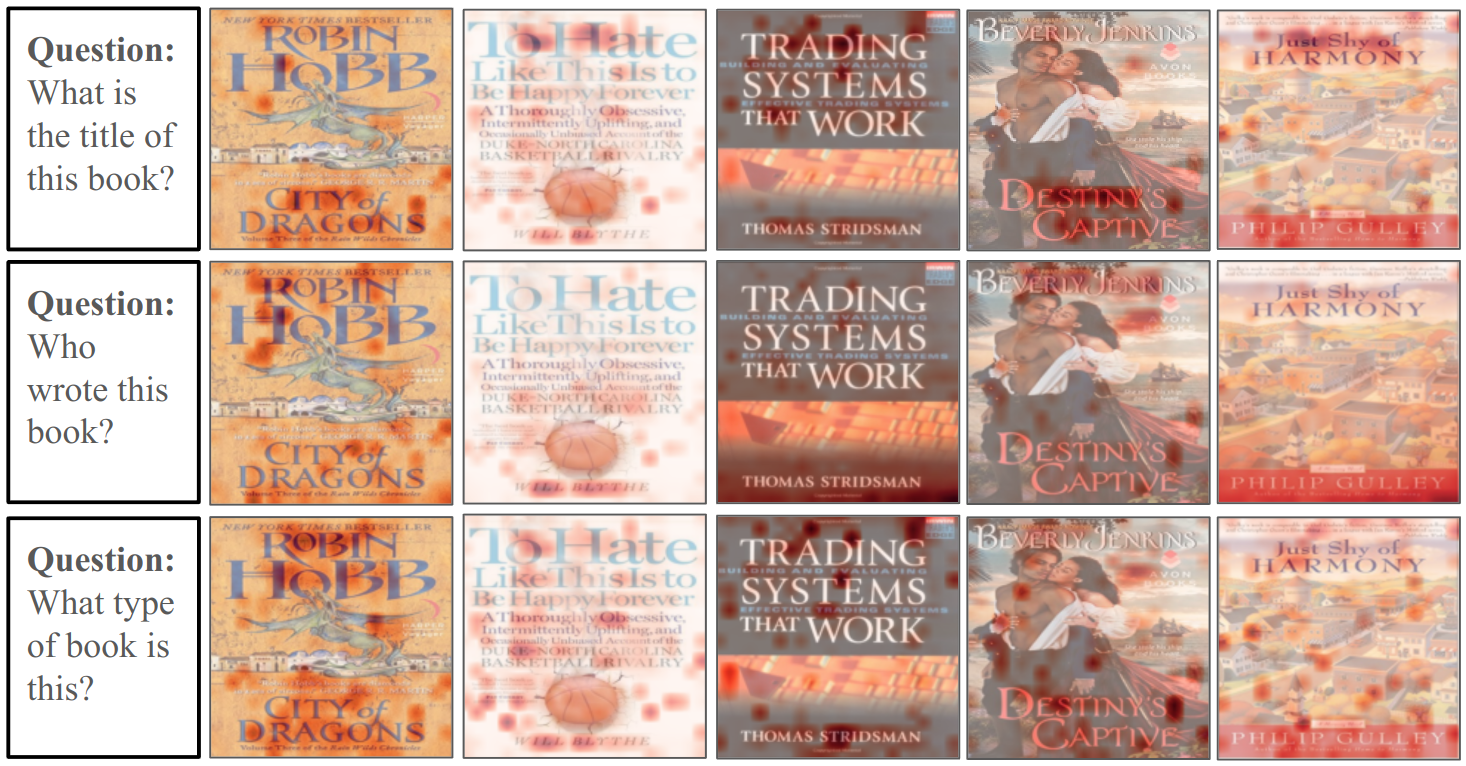}
        }\vspace{2pt}\\
        \subfloat{
            \includegraphics[width=\columnwidth]{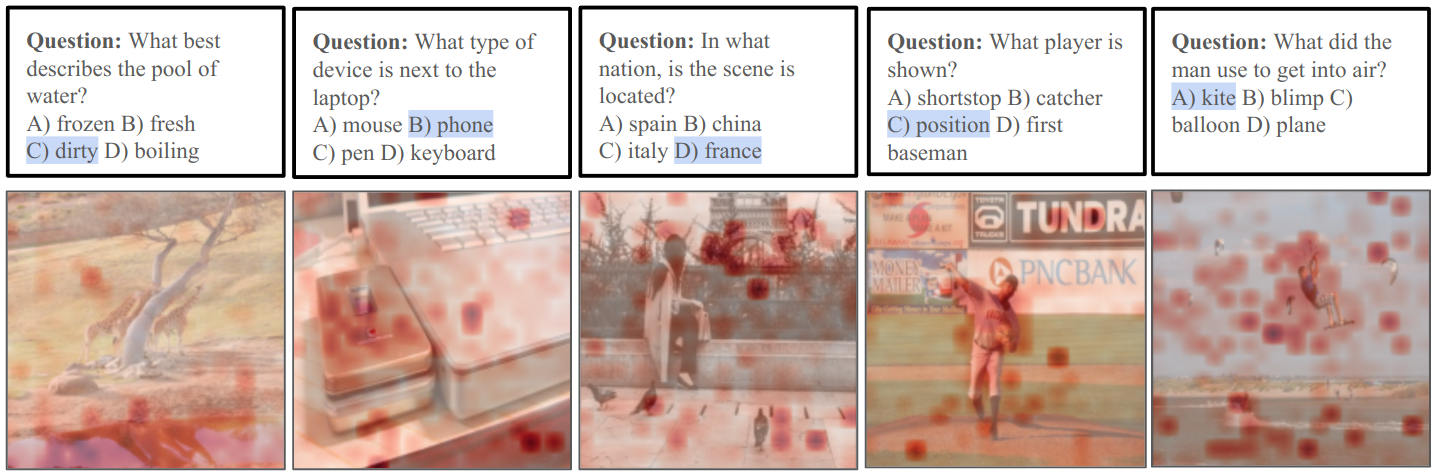}%
        }
        \caption{Visualization of tracked attention-score statistics (Eq.~\ref{eq:imp}) on various image QA pairs from OCR-VQA and A-OKVQA.}
        \vspace{-15pt}
        \label{fig:examples}
\end{figure}

\subsection{Compatibility with Other Techniques}

\textbf{Integration with Eviction and Quantization:} To analyze the potential of our technique, we perform experiments combining AttentionPack with eviction and also quantization. It is important to explore the complementary behavior of technique, especially for environments with strict memory restriction that may need aggressive cache size reduction. To this end, we first replace attention-aware decompression with eviction such that the tokens determined to be decompressed with lower rank based on Eq.~\ref{eq:imp} are evicted. We also analyze the performance in combination with 4-bit quantization following KVQuant~\cite{kvquant}. As reported in Table~\ref{tab:quant}, we observe that AttentionPack can achieve further gains when combined with quantization and eviction. We see around 0.5\% performance drop compared to KVQuant when we combine quantization with our approach while having $~5\times$ smaller cache and $2\times$ faster inference. Eviction affects the performance negatively in OCR-VQA as the visual tokens needed for accurate answer may change at each step.

\noindent
\textbf{Integration with Grouped Query Attention: } Groped query attention (GQA) is a variation of the attention mechanism used in transformer models, designed to improve efficiency and scalability. GQA reduces the computational cost by having fewer key-value groups than query heads. To analyze the compatibility of our approach with GQA, we perform experiments with the recent video-language model Qwen3VL-8B-instruct, which utilizes this technique. In Table~\ref{tab:qwen3}, we also report the results obtained on MMMU with competitive performance at 3.17x smaller cache.

\noindent
\textbf{Integration with Lower-level Optimizations: } To achieve further latency reductions, we also demonstrate that integration with low-level optimizations such as FlashAttention~\cite{flashattention} is also possible. To this end, we provide a fused kernel implementation, which merges the decompression operation directly with the attention score computation (Algorithm~\ref{alg_1}, lines 7-12). The principle is similar to FlashAttention~\cite{flashattention}, which optimizes I/O by reducing data transfers between HBM and SRAM. We apply this idea by incorporating our decompression step into FlashAttention-v1. This approach successfully reduces latency while still permitting the materialization of attention scores for our importance estimation. Fused kernel optimizes memory access by for each query, tiling the key and value cache blocks along the sequential axis such that the decompression and multiplication happens on-the-fly in SRAM together just before the attention score calculation with online softmax. Empirical results on LLaVA1.5-7B (Fig~\ref{fig:kernel_latency}) confirm the benefits, where the fused kernel almost halves the decoding latency for 32 tokens in both single and batch inference settings when compared to our standard implementation.

\subsection{Qualitative Analysis}

To understand how the attention score statistics utilized in attention-aware decompression, we illustrate several image-question pairs  from MSVD-QA in Figure~\ref{fig:examples_video} and OCR-VQA, A-OKVQA in Figure~\ref{fig:examples}. We observe that the tracked importance scores highlights various regions of the visual input depending on the question and not every visual token is equally used at each decoding step. The model tends to put less attention on backgrounds and more on the objects related to the desired answer. These examples show the validity of the motivation behind our attention-aware decompression technique, which reduces decompression overhead without hurting output quality.

\section{Conclusion}
\label{sec:conclusion}

We have proposed AttentionPack, an attention-aware memory-efficiency optimization framework for decoding with large vision-language models. Our approach introduces two key technical contributions: (1) a multi-head compression technique leveraging singular value decomposition to reduce the size of stored key and value vectors, and (2) an attention-aware partial decompression technique, minimizing latency overhead and addressing memory constraints in long-context inference. Thanks to reduced memory usage, AttentionPack enables larger batch sizes and extended context lengths while preserving the model output quality and performance. Comprehensive evaluations on five image and video QA datasets demonstrate that our approach can reduce the key-value cache size by up to 8 times and improve batch inference throughput by up to 50\%.

\vspace{12pt}
\noindent {\bf Acknowledgement.\/} 
This research is partially sponsored by NSF CISE grants 2302720 and 2312758, an IBM faculty award, a grant from the CISCO Edge AI program, and the research cyberinfrastructure resources and services provided by the Partnership for an Advanced Computing Environment (PACE) at the Georgia Institute of Technology. Fatih Ilhan and Ling Liu are the primary contact authors for this work. 

{
    \small
    \bibliographystyle{ieeenat_fullname}
    \bibliography{main}
}

\clearpage

\section{Algorithmic Description}
\label{sec:algo}
In this section, we outline the operations within a single decoding step of an LVLM, containing the operations for compression, attention-aware decompression, and attention score computations, as pseudocode in Algorithm~\autoref{alg_1}. For simplicity, we omit subscripts differentiating visual and textual tokens in algorithm statements, as the same procedure applies to both. Before compression is performed (i.e., when the compressed cache is empty, \( T_{cc} = 0 \)), we have the uncompressed cache \( \mathcal{C}_u = \{\mathbf{K}, \mathbf{V} \in \mathbb{R}^{T_{uc} \times HD} \} \), storing the key/value vectors for all \( T_{uc} \) tokens from the prefill phase. After computing the query \( \boldsymbol{q} \), key \( \boldsymbol{k} \), and value \( \boldsymbol{v} \) vectors for the current input (line 3), these key/vectors are appended to the uncompressed cache (line 4). If compression had already been performed (\( T_{cc} > 0 \)), tokens are decompressed using partial ranks derived based on the attention score statistics (lines 5-6). These decompressed tokens are then merged with the uncompressed cache (lines 7-10). The attention scores (line 11) and output (line 12) are computed, and the attention score statistics are updated (line 13). At any decoding step, if the uncompressed cache length ($T_{uc}$) exceeds the compression period \( T_p \), compression is triggered (lines 14-17) to generate the compressed cache and decompression matrices. Once compression is done, the uncompressed cache is emptied and begins collecting new tokens in next decoding steps until it reaches \( T_p \) again, when the compression is reapplied. Finally, the cached token statistics \( \mathcal{I} \), the output \( \mathbf{O} \), the compressed cache \( \mathcal{C}_c \), and the updated uncompressed cache \( \mathcal{C}_u \) are returned for use in future decoding steps.

\begin{algorithm}[h!]
\caption{\small\texttt{Attention Block Operations in Decoding Step-$t$}}
    \begin{algorithmic}[1]
        {\small
            \STATE \textbf{Inputs:} input activation $\mathbf{H} \in \mathbb{R}^{T_q \times HD}$, uncompressed cache for visual ($*=v$) and textual ($*=t$) tokens $\mathcal{C}_{u*} = \{\mathbf{K}_{*}, \mathbf{V}_{*} \in \mathbb{R}^{T_{uc*}\times HD}\}$, compressed cache and decompression matrices
            $\mathcal{C}_{c*} = \{\overline{\mathbf{K}}_{*} \in \mathbb{R}^{T_{cc*}\times R_{k*}}, \overline{\mathbf{V}}_{*} \in \mathbb{R}^{T_{cc*}\times R_{v*}}, \mathbf{D}_{k*} \in \mathbb{R}^{R_{k*} \times HD}, \mathbf{D}_v \in \mathbb{R}^{R_{v*} \times HD}\}$, cached token importance scores $\mathcal{I}$, attention block weights
            $\boldsymbol{\theta} = \{\mathbf{W}_q, \mathbf{W}_k, \mathbf{W}_v, \mathbf{W}_o \in \mathbb{R}^{HD \times HD}\}$
            \STATE \textbf{Parameters:} $T_{p}$ (compression period), $F$ (number of decompression groups), $R_{k*}, R_{v*}$ (compression rank for keys and values)
            \STATE Compute new query, key, value vectors: $\mathbf{Q'}, \mathbf{K'}, \mathbf{V'} \leftarrow \mathbf{H}\mathbf{W}_{\dagger}$ for $\dagger \in \{q, k, v\}$
            \STATE Append new key, value vectors to corresponding uncompressed cache: $\mathbf{K} \leftarrow \mathbf{K} \oplus \mathbf{K'}$, $\mathbf{V}_{*} \leftarrow \mathbf{V} \oplus \mathbf{V'}$\noindent
            \IF{$T_{cc} > 0$}
                \STATE Determine the tokens $\boldsymbol{m}_{f}$ for each group $f\in\{1, ..., F\}$ based on $\mathcal{I}$
                \STATE Retrieve cache: $\mathbf{\Tilde{K}} \leftarrow \bigoplus_{f=1}^F \overline{\mathbf{K}}[\boldsymbol{m}_{f}, :R_k^{(f)}]\mathbf{D}_k[:R_k^{(f)}] \oplus \mathbf{K}$; $\mathbf{\Tilde{V}} \leftarrow  \bigoplus_{f=1}^F\overline{\mathbf{V}}[\boldsymbol{m}_{f}, :R_v^{(f)}]\mathbf{D}_v[:R_v^{(f)}] \oplus \mathbf{V}$
            \ELSE
                \STATE Retrieve cache: $\mathbf{\Tilde{K}} \leftarrow \mathbf{K}$; $\mathbf{\Tilde{V}} \leftarrow \mathbf{V}$
            \ENDIF
            \STATE Compute attention scores: $\mathbf{A} \leftarrow \texttt{softmax}(\frac{\mathbf{Q}\mathbf{\Tilde{K}}^T}{\sqrt{D}})$
            \STATE Compute output: $\mathbf{O} \leftarrow (\mathbf{A}\mathbf{\Tilde{V}})\mathbf{W}_{o}$
            \STATE Update statistics $\mathcal{I}$ based on $\mathbf{A}$ using Eq.~\ref{eq:imp}.
            \IF{$T_{uc} \geq T_{p}$}
                \STATE Compress cache: $\overline{\mathbf{K}}, \mathbf{D}_{k} \leftarrow \texttt{svd}(\mathbf{\Tilde{K}}, R_{k})$; $\overline{\mathbf{V}}, \mathbf{D}_{v} \leftarrow \texttt{svd}(\mathbf{\Tilde{V}}, R_{v})$\noindent
                \STATE Delete $\mathbf{K}$, $\mathbf{V}$
            \ENDIF
            \STATE \textbf{return} $\mathbf{O}$, $\mathcal{C}_{u}$, $\mathcal{C}_{c}$, $\mathcal{I}$
        }
	\end{algorithmic} \label{alg_1}
\end{algorithm}

\section{Experiment Setup Details}
\label{sec:app_setup}
\subsection{Datasets}

We experiment on five datasets from different image/video tasks and domains. First, we consider three question-answering datasets based on image understanding: (i) A-OKVQA~\cite{aokvqa}, a multiple-choice question dataset requiring a broad base of commonsense and world knowledge containing 1147 test images, (ii) OCR-VQA~\cite{ocrvqa}, a visual understanding dataset containing book cover images and related questions requiring optical character recognition capabilities containing 2382 test images, (iii) MMMU~\cite{mmmu} a multi-discipline multi-modal
reasoning benchmark containing questions from college exams containing 900 test images. Second, we consider two video understanding benchmarks: (i) MSVD-QA~\cite{msvdqa} and (ii) MSRVTT-QA~\cite{msvdqa} containing short video clips with multiple question-answer pairs for each video, with 1970 and 2990 test videos, respectively.

\subsubsection{Implementation Details}

For image QA tasks, we experiment with 7B and 13B variants of the LLaVA1.5~\cite{llava}, which is an end-to-end trained large multi-modal model that connects a visual encoder backbone (CLIP~\cite{clip}) with an LLM decoder (Vicuna~\cite{vicuna}). LLaVA1.5 converts each input image to 24x24 representation, resulting with 576 tokens. We also consider QwenVL-Chat-7B for experiments with grouped query attention~\cite{qwen}. In video QA tasks, we consider VideoLLaVA-7B, which has a similar architecture with LLaVA, but processes visual input through image and video encoders~\cite{languagebind}, and pre-aligns before feeding into the LLM decoder. VideoLLaVA converts each frame to 14x14 representation, resulting in 256 tokens per frame. The code is available at \href{https://github.com/git-disl/AttentionPack}{https://github.com/git-disl/AttentionPack}.

\begin{table*}[t!]
\renewcommand{\arraystretch}{0.77}
\begin{adjustbox}{width=\textwidth}
\begin{tabular}{@{}lcccccccc@{}}
\toprule
\multirow{3}{*}{\textbf{Method}} & \multirow{3}{*}{\begin{tabular}[c]{@{}c@{}}\textbf{Average Cache Size}\\ \textbf{per Instance (MB)} \end{tabular}} & \multicolumn{7}{c}{\textbf{Dataset}} \\
& & \multicolumn{4}{c}{\begin{tabular}[c]{@{}c@{}}\textit{Question Answering - F1 (\%)} \end{tabular}} & \multicolumn{2}{c}{\begin{tabular}[c]{@{}c@{}}\textit{Summarization - ROUGE-L (\%)} \end{tabular}} & \multirow{2}{*}{avg.} \\
& & qasper & mfqa\_en & hotpotqa & 2wikimqa & gov\_report & multi\_news & \\ \midrule
Full KV & $555.36$ & $34.68$ & $42.66$ & $43.66$ & $35.65$ & $29.27$ & $26.68$ & $35.43$\\
\cdashline{1-9}\noalign{\vskip 0.4ex}
H2O ($T=1000$) & $131.30$ & ${31.68}$ & $40.35$ & $42.09$ & $34.44$ & $\bm{27.64}$ & $25.03$ & $33.53$\\
\textbf{AttentionPack ($T=1500$)} & ${{119.39}}$ & $\bm{33.86}$ & $\bm{41.00}$ & $\bm{42.47}$ & $\bm{34.58}$ & $27.26$ & $\bm{25.67}$ & $\bm{34.14}$\\
\textbf{AttentionPack ($T=1000$)} & ${{76.71}}$ & $30.60$ & $39.71$ & $42.10$ & $33.89$ & $25.99$ & $24.82$ & $32.85$\\
 \bottomrule
\end{tabular}
\end{adjustbox}
\caption{Text QA and summarization results on LongBench datasets with LLaMA3.1-8B.}
\label{tab:longbench}
\end{table*}

For comparisons, we consider standard inference with full KV caching and two recent token eviction techniques developed for LLMs: H2O~\cite{h2o}, Scissorhands~\cite{scissor}, a token skipping technique for LVLMS: FastV~\cite{fastv} and a recent cache compression technique Minicache~\cite{mini}. For our approach, during compression, we use randomized SVD~\cite{random_svd} for key and value compression with $R_{kv}=R_{vv}\in\{32,64\}$ for visual tokens in image QA, $R_{kv}=R_{vv}=\in\{64,128\}$ in video QA and do not apply compression for textual tokens, unless stated otherwise. We perform attention-aware decompression over value cache with $F=2, r_1=0.25, r_2=0.75, R_{vv}=R_{vv}^{(1)}=4R_{vv}^{(1)}$. In initial experiments, $\alpha \in [0.05, 0.75]$ gave robust and stable results and we set $\alpha=0.25$. We use half-precision except for the attention computation, which is at full-precision.

\section{Additional Results and Analysis}
\label{sec:app_results}

\subsection{Latency at Various Context Lengths}
\label{sec:app_latency}

\begin{figure}[t!]
    \centering
    \includegraphics[width=\columnwidth]{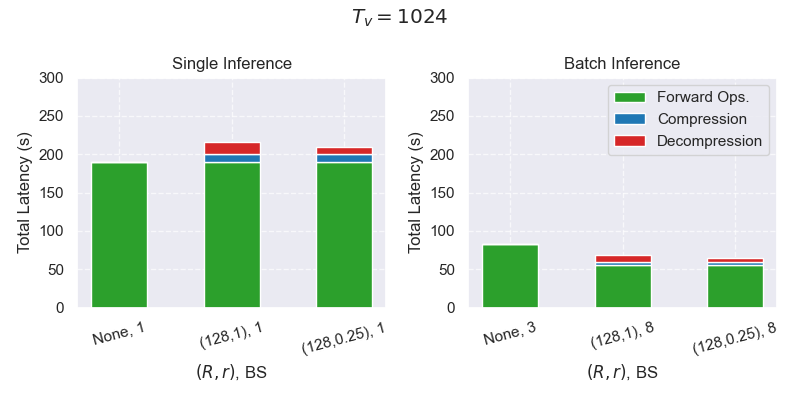} \\
    \vspace{-10pt}
    \includegraphics[width=\columnwidth]{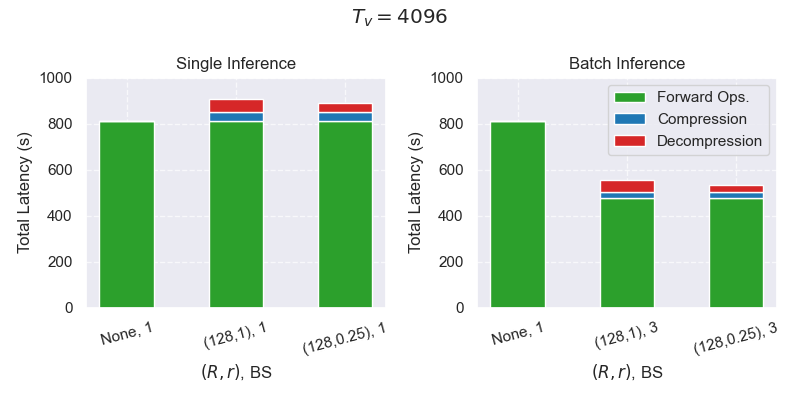}
    \vspace{-15pt}
    \caption{Total decoding latency for 100 queries across various context lengths (4/16 frames resulting in 1024/4096 visual tokens).}
    \vspace{-5pt}
    \label{fig:latency_}
\end{figure}

For latency analysis at various context lengths, we repeat the setting in Figure~\ref{fig:latency} for MSVD-QA using VideoLLaVA-7B with number of frames in $\{4, 16\}$ resulting in $\{1024, 4096\}$ visual tokens. We observe latency gains in batch inference are valid across various context lengths as shown in Figure~\ref{fig:latency_}.

\subsection{Analysis of Textual Token Compression}

\begin{figure}[t!]
\centering
    \includegraphics[width=\columnwidth]{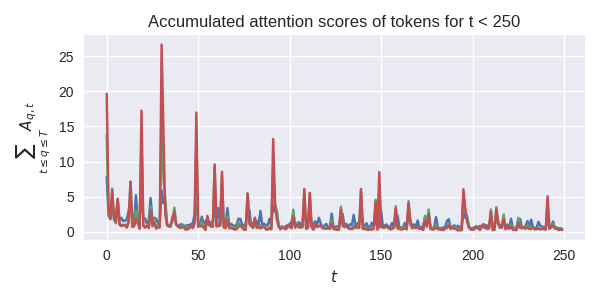}
    \caption{Accumulated attention scores of the first 250 tokens at the timestep $T=500$ from a random conversational text sample from MTBench~\cite{mtbench} (for the first three layers, which are represented with different colors).}
    \label{fig:score_plots}
\end{figure}

Given the vision-language datasets considered have most of the context from visual inputs, e.g., the number of visual tokens in image QA is 576, and textual tokens is less than 100; treating visual tokens with a rank $R << min(T, HD)$ for high compression offers better cache size efficiency. In initial experiments, we observed compressing the visual and textual tokens together resulted in very poor performance (around 20\% loss in A-OKVQA). In contrast, focusing on high compression ratio of visual tokens, though simple, shows big gains. Still, to analyze the text domain effectiveness, we conducted a small experiment with LLaMA3.1-8B~\cite{llama3} on LongBench~\cite{longbench} datasets and compared with the representative cache eviction technique~\cite{h2o}. Cache length, avg. cache size (MB) and F1 scores on question-answering, ROUGE-L on summarization tasks are reported. We observe that value matrices requires higher rank compared to key matrices in these datasets, which motivates us to only compress key matrices ($R_k=64$). Our approach is effective, with around 40\% cache reduction under same cache length, and +0.6\% performance under same cache size. H2O is effective by evicting tokens with low accumulated attention scores as shown in Figure~\ref{fig:score_plots}. But, AttentionPack enables longer context windows thanks to compression along hidden dimension and yields better performance under same constraints.

\subsection{Impact of Multi-Head Compression} 
We carry out an ablation study on OCR-VQA to validate the analysis in Figure~\ref{fig:lr}. For OCR-VQA, we find that achieving similar performance (52.25\%) requires  1.48x larger cache (3.44x comp. ratio) when heads are compressed separately with $R_{kv}=24$. Under the same compression ratio (5x), multi-head approach outperforms by 1.8\%.

\end{document}